%% file: main.tex
\documentclass[runningheads]{llncs}
\usepackage[T1]{fontenc}
\usepackage{hyperref}
\usepackage{url}
\usepackage{graphicx,verbatim}

\usepackage{booktabs}
\usepackage{pifont}
\newcommand{\cmark}{\ding{51}}
\newcommand{\xmark}{\ding{55}}
\usepackage{xcolor}
\usepackage{amsmath}
\usepackage{multirow}
\usepackage{makecell}
\usepackage{booktabs} 
\usepackage{pifont}   
\usepackage{xcolor}   

\newcommand{\tick}{\textcolor{green!70!black}{\ding{51}}} 
\newcommand{\cross}{\textcolor{red}{\ding{55}}}          
\begin{document}
\title{
SurgOnAir: Hierarchy-Aware Real-Time Surgical Video Commentary
}

\author{Jingyi He$^{*}$\inst{1} \and 
Yue Zhou$^{*}$\inst{1,2} \and 
Long Bai\inst{4} \and \\ 
Kun Yuan\inst{1,2,3} \and 
Nassir Navab\inst{1,2} \and 
Yuan Bi\inst{1,2}}

\institute{
Computer Aided Medical Procedures (CAMP),\\ TU Munich, Germany \\
\email{kun.yuan@ext.ihu-Strasbourg.eu} \\
\and Munich Center for Machine Learning (MCML), Munich, Germany 
\and University of Strasbourg, France \\
\and The Chinese University of Hong Kong, Hong Kong
}

\renewcommand{\thefootnote}{\fnsymbol{footnote}}
\footnotetext[1]{Equal contribution.}
\authorrunning{He et al.}
\maketitle              
\input{sections/0_Abstract}
\input{sections/1_Introduction}

\input{sections/2_Method}

\input{sections/3_Experiment}

\input{sections/4_Conclusion}

\bibliographystyle{splncs04}
\bibliography{reference}

\end{document}

%% file: sections/0_Abstract.tex
\begin{abstract}
Understanding surgical workflow in real time is fundamental for intelligent surgical embodiment, where AI systems continuously perceive and respond as surgery proceeds. In the operating room, critical decisions depend on subtle, moment-to-moment changes, such as fine instrument movements and evolving tissue states, where even slight perceptual delays can limit assistance or compromise safety. Yet existing methods remain offline or operate at coarse temporal scales, generating descriptions only after processing clips, preventing immediate reaction. We address this by proposing SurgOnAir, a streaming vision-language model that processes frames sequentially without future access and progressively generates narration tokens as visual input arrives. 
SurgOnAir achieves fine-grained frame-to-token generation, enabling instant responsiveness to evolving surgical dynamics. 
Built upon our curated hierarchical dataset SurgOnAir-11k spanning action-, step-, and phase-level supervision, the model is trained to produce multi-level textual responses that reflect the inherent hierarchy of surgical procedures. Furthermore, special transition tokens are generated to explicitly mark state changes, allowing SurgOnAir to capture and signal key workflow transitions as they occur. 
Experiments show that SurgOnAir enables real-time understanding through a single vision-language model that unifies streaming across multiple hierarchies of the surgical workflow, generating superior and hierarchy-aware narrations. The project is publicly available at \url{https://github.com/KarolineZhy/SurgOnAir}.
\end{abstract}

%% file: sections/1_Introduction.tex
\section{Introduction}
Surgical procedures are inherently dynamic, evolving through continuous and multi-stage transitions. As surgical robotics advances toward higher autonomy, the surgical AI systems must develop temporally grounded understanding to perceive the environment and make informed real-time decisions~\cite{long2025surgical,kim2025srt,wei2026moves}.Recent multimodal large language models (MLLMs)~\cite{bai2025qwen3,jiang2025hulu} have significantly advanced multimodal perception and reasoning in general domains, while their potential is also being increasingly explored in medical applications. These efforts span a broad range of clinical scenarios, including but not limited to ultrasound imaging~\cite{zhou2025ultraad,zhou2026look,bi2026scanning}, surgical videos~\cite{wang2025surgvidlm,li2024llava}, and X-ray imaging~\cite{li2025enhancing}.

In the natural video domain, MLLMs have achieved remarkable success in comprehending complex spatial-temporal sequences, as evidenced by pioneering frameworks like Video-ChatGPT~\cite{maaz2024video}, Chat-UniVi~\cite{jin2024chat}, and Video-LLaVA~\cite{lin2024video}. 
Driven by these breakthroughs, researchers have rapidly migrated such paradigms into the surgical domain for understanding the dynamics of surgical scenes. LLaVA-Surg~\cite{li2024llava} curates a large-scale dataset to train a multimodal conversational assistant for surgical video question-answering. Later, SurgVidLM~\cite{wang2025surgvidlm} addresses both holistic and fine-grained surgical video comprehension by proposing a two-stage mechanism that fuses global procedural context with local details. 
Despite their strong performance, most existing surgical video understanding methods are designed under an offline setting. They assume access to the entire video sequence and perform captioning or question answering retrospectively, limiting their applicability in real-time surgical scenarios.

\begin{table}[t]
\centering
\caption{Comparison of surgical workflow analysis models in hierarchical and streaming capability.}
\label{tab:comparison}

\setlength{\tabcolsep}{3pt}
\renewcommand{\arraystretch}{0.92}

\begin{tabular}{l|ccc|cc}
\toprule
& \multicolumn{3}{c|}{\textbf{Hierarchy}} 
& \multicolumn{2}{c}{\textbf{Caption}} \\
\cmidrule(lr){2-4} \cmidrule(lr){5-6}
\textbf{Dataset}
& Phase & Step & Action 
& Offline & Stream \\
\midrule
EndoNet~\cite{endonet} 
& \tick & \cross & \cross 
& \cross & \cross \\
Gastric Bypass~\cite{Lavanchy2024}  
& \tick & \tick & \cross  
& \cross & \cross \\
Rendezvous~\cite{nwoye2023cholectriplet2021} 
& \cross & \cross & \tick 
& \cross & \cross \\
DAISI~\cite{rojasmuñoz2020daisidatabaseaisurgical} 
& \cross & \cross & \cross 
& \tick & \cross \\
HecVL~\cite{yuan2024hecvl} 
& \tick & \tick & \tick 
& \cross & \cross \\
LLaVA-Surg~\cite{li2024llava} 
& \cross & \cross & \cross 
& \tick & \cross \\
SurgVidLM~\cite{wang2025surgvidlm} 
& \tick & \tick & \tick 
& \cross & \cross \\
\midrule
\textbf{Ours} 
& \tick & \tick & \tick 
& \tick & \tick \\
\bottomrule
\end{tabular}
\end{table}

Recent efforts have begun exploring online video understanding.
VideoLLM-Online~\cite{chen2024videollm} introduces a Learning-In-Video-Stream framework that models video understanding as continuous streaming dialogue, enabling temporally aligned real-time interaction.
Building on this direction, LiveCC~\cite{chen2025livecc} proposes a timestamp-aligned streaming training scheme that densely interleaves Automatic Speech Recognition (ASR) words with video frames, facilitating fine-grained speech-vision alignment.
More recently, StreamingVLM~\cite{xu2025streamingvlm} focuses on memory-efficient inference for near-infinite video streams via a compact KV-cache. 
However, modeling them as flat sequences overlooks the inherent hierarchical organization which is essential for surgical procedures understanding since surgical workflow is inherently hierarchical, spanning phases, steps, and actions~\cite{wang2025beyond}. 

Hierarchical modeling is crucial in surgical workflows, as hierarchical structure provides contextual grounding that improves cross-context disambiguation and reduces hallucination. HecVL~\cite{yuan2024hecvl} introduces a hierarchical video-language pretraining framework that pairs surgical videos with multi-level textual supervision, and learns disentangled representations across fine-to-coarse hierarchies. Similarly, PeskaVLP~\cite{yuan2024procedure} proposes a hierarchical knowledge-augmented pretraining strategy that integrates refined textual supervision with dynamic temporal alignment for workflow-aware representation learning. However, these approaches focus on hierarchical representation learning in offline pretraining settings and are not formulated as generative, streaming models for surgical video narration. However, these approaches use hierarchy for discriminative workflow recognition, whereas SurgOnAir incorporates this into generative streaming narration to maintain contextual consistency during long-horizon generation.

Therefore, it is essential to design a generative live-streaming narration model that explicitly incorporates hierarchical structure inherent in surgical procedures for narration. As illustrated in Tab.~\ref{tab:comparison}, existing surgical understanding models lack either the support to complete Phase-Step-Action hierarchy or for both offline and streaming captioning. To this end, we introduce SurgOnAir, a hierarchy-aware streaming surgical narration model capable of real-time generation. To achieve this, we propose the SurgOnAir-11K, a hierarchically paired video-language dataset, where surgical clips are aligned with temporally grounded narration at the word level and enriched with temporally annotated hierarchical phase and step labels. This design enables temporally aligned hierarchical supervision for live-streaming surgical narration. By incorporating hierarchical supervision as structured state constraints, SurgOnAir learns to capture workflow transitions and cross-granularity dynamics across phases, steps, and actions.

Our contributions are threefold:
(1) We design a hierarchical temporal grounding pipeline that converts raw surgical videos and ASR transcripts into multi-level, temporally aligned video-language supervision, yielding SurgOnAir-11K.
(2) We introduce SurgOnAir, a hierarchical streaming framework which interleaves multi-modal visual-text tokens for real-time processing, while leveraging specialized transition tokens to explicitly predict state changes for hierarchical-aware narration. 
(3) Extensive experiments demonstrate that our approach  outperforms both existing offline and streaming models, achieving superior real-time narration quality and enhanced hierarchical awareness at critical procedural transitions.

%% file: sections/2_Method.tex
\section{Method}

\subsection{Hierarchical Temporal Grounding Data Curation Pipeline}
\begin{figure}[t]
    \centering
    \includegraphics[width=\linewidth]{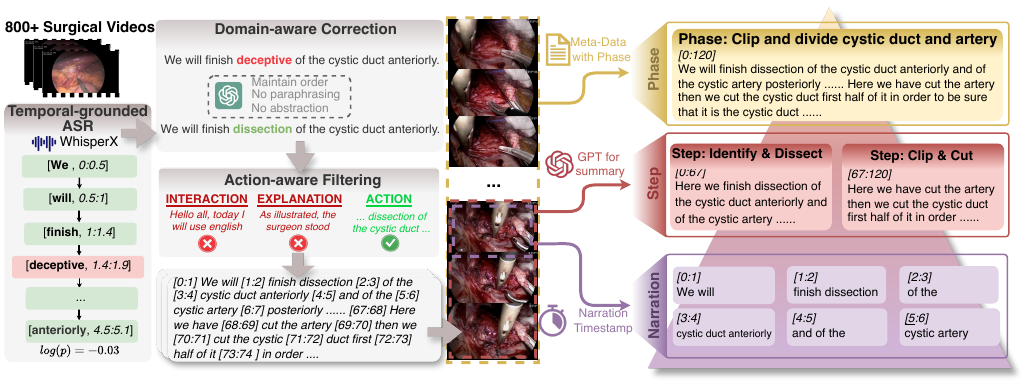}
    \caption{Overview of the hierarchical temporal grounding pipeline for narration data curation.} 
    \label{fig:datacuration}
\end{figure}

\noindent
\textbf{Overview.}
To generate visually-grounded video-language supervision, we introduce an automatic pipeline (Fig.~\ref{fig:datacuration}). We detail each module below.

\noindent
\textbf{Temporally Grounded ASR.} We use WhisperX-large~\cite{bain2023whisperx} to obtain word-level transcripts with per-word start/end times and a per-sentence average log-probability $\log(p)$.

\noindent
\textbf{Domain-aware ASR Correction.}
Sentences with $\log(p) < -0.15$ (probability $< 0.865$) are flagged as low-confidence and rewritten with GPT-4o~\cite{achiam2023gpt} under a strict constraint: only terminology normalization and recognition-error correction, with no content added or removed.

\noindent
\textbf{Action-aware Filtering.}
As surgical narration is discursively heterogeneous, we prompt GPT-4o~\cite{achiam2023gpt} to label each sentence as \textit{ACTION}, \textit{EXPLANATION}, or \textit{INTERACTION}, and keep only \textit{ACTION} sentences for supervision, since the rest lack direct alignment with the video frames (Fig.~\ref{fig:datacuration}).

\noindent
\textbf{Hierarchical Data Construction.}
We organize the data into a three-level Phase~$\rightarrow$~Step~$\rightarrow$~Word hierarchy. Word-level \textit{ACTION} sentences provide temporal grounding; an LLM groups them into 1--3 high-level steps per phase as a monotonic, non-overlapping partition; these steps are then aligned with phase definitions and temporal boundaries from the video metadata.

\subsection{SurgOnAir Streams Surgical Workflows Hierarchically}

\begin{figure}[t]
    \centering
    \includegraphics[width=\linewidth]{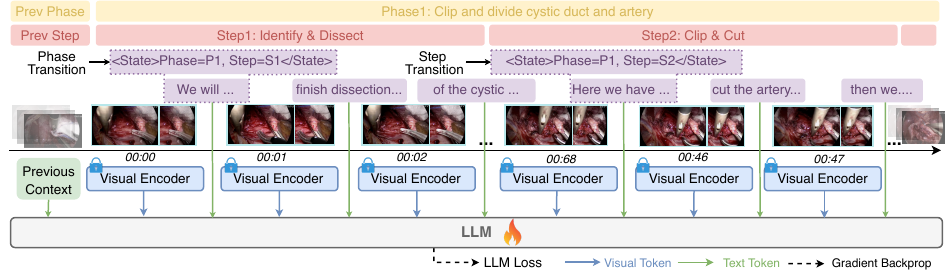}
    \caption{\textbf{Overview of our hierarchical streaming training architecture.} Visual tokens extracted from temporally aligned frames are interleaved with narration tokens in a causal LLM. Structured  \texttt{<State>} tokens explicitly encode Phase and Step states, marking hierarchical state transitions along the timeline.}
    \label{fig:pipeline}
\end{figure}

To effectively synchronize the stream of visual frames with real-time speech, we adopt the dense temporal interleaving strategy introduced in~\cite{chen2025livecc}. 
The interleaved input sequence is formulated as:
\begin{equation}
\begin{aligned}
[\mathrm{Con}] \; \langle F_{t:t+k} \rangle \; \langle W_{t:t+k} \rangle \; \langle F_{t+k:t+2k} \rangle \; \langle W_{t+k:t+2k} \rangle \; \\
\dots \; \langle F_{t+n k:t+(n+1)k} \rangle \; \langle W_{t+n k:t+(n+1)k} \rangle \;,
\end{aligned}
\label{eq:multimodal_input}
\end{equation}
where $[\mathrm{Con}]$ denotes the preceding contextual metadata (e.g., initial prompts or historical context shown as ``Previous Context'' in Fig.~\ref{fig:pipeline}), $\langle F \rangle$ denotes visual frames extracted by the visual encoder, $\langle W \rangle$ represents ASR and generated text tokens, $t$ is the temporal index, and $k$ is the stride. 

To enforce hierarchy-aware surgical narration, we incorporate hierarchical procedural supervision via special \texttt{<State>} tokens. Specifically, the text sequence $\langle W_{t:t+k} \rangle$ is not merely a flat stream of words, instead it dynamically encapsulates both the continuous procedure narration and the discrete structural workflow boundaries. We formulate $\langle W_{t:t+k} \rangle$ as a conditional concatenation:
\begin{equation}
    \langle W_{t:t+k} \rangle = 
    \begin{cases} 
        \langle S_{t} \rangle \oplus \langle N_{t:t+k} \rangle, & \text{if a state transition occurs within } [t, t+k) \\ 
        \langle N_{t:t+k} \rangle, & \text{otherwise} 
    \end{cases}
\label{eq:text_sequence}
\end{equation}
where $\oplus$ denotes sequence concatenation, and $\langle N_{t:t+k} \rangle$ represents the continuous text narration. The state sequence $\langle S_{t} \rangle$ acts as a discrete workflow anchor (e.g., \texttt{<State>Phase=P1, Step=S1</State>}), which provides hierarchical context around phase or step transitions. This dual-level formulation allows the model to perform continuous narration while remaining aware of the underlying workflow structure.

\noindent
\textbf{Training Details} 
We adopt the pre-trained LiveCC-7B-Base~\cite{chen2025livecc} with streaming capability as the backbone. During fine-tuning, only the LLM is updated. To enable fine-grained alignment and reduce redundancy, we set the stride $k=1$ and the frame rate to 2 FPS. The model is trained with a learning rate of $2e-5$. To enhance visual grounding and reduce over-reliance on textual context, we adopt a staged conditioning strategy: in the first 2 epochs, the model is conditioned on previous ASR text to stabilize training and maintain temporal coherence. In the final epoch, the ASR context is removed and only the video title is provided, forcing the model to rely more directly on visual evidence.

%% file: sections/3_Experiment.tex
\section{Experiment}
\textbf{Dataset}
We employ stratification of SurgOnAir-11K into training and testing split by surgical meta-type to ensure a representative distribution of diverse procedural categories and maintain balanced complexity across both sets. 

\noindent
\textbf{Baseline}
We evaluate our method against two categories of baselines: offline Video LLMs, including LLaVA-Video-7B~\cite{lin2024video} and Qwen2.5-VL-7B~\cite{bai2025qwen3}, and the streaming-based LiveCC-7B~\cite{chen2025livecc}. While the offline models follow a conventional captioning protocol by processing all frames simultaneously, LiveCC-7B serves as our primary competitive baseline for real-time, frame-by-frame generation. Notably, during evaluation, we provide only the video title as context to force the model to rely on salient visual cues rather than textual continuation from prior ASR transcripts.

\noindent
\textbf{Evaluation Protocal}
We formulate evaluation as a pairwise choice task for GPT-4o~\cite{achiam2023gpt}, acting as an LLM-as-a-judge. Conditioned on the ground-truth ASR transcripts, the judge is prompted to explicitly select the superior narration between two models. We report the win rate~\cite{chen2025livecc}, defined as the percentage of wins against the other model. To align with the practical requirements of streaming surgical narration, the evaluation paradigm prioritizes semantic correctness, specifically focusing on the precise grounding of anatomical and procedural content rather than syntactic perfection.

\begin{table}[t]
\centering
\small

\begin{minipage}[t]{0.45\textwidth}
\centering
\caption{Comparison of offline and streaming models for captioning}
\label{tab:model_comparison_a}
\begin{tabular}{l|c|c}
\toprule
Model & Live? & Win Rate \\
\midrule
Hulu-Med-7B~\cite{jiang2025hulu}      & \xmark & \textcolor{red}{\ding{96}} \\
LLaVA-Video-7B~\cite{lin2024video}   & \xmark & 11.3 \\
Qwen2.5-VL-7B~\cite{bai2025qwen3} & \xmark & 6.2  \\
\midrule
LiveCC-7B~\cite{chen2025livecc}  & \cmark & 16.7 \\
SurgOnAir-base       & \cmark & 60.4 \\
SurgOnAir       & \cmark & \textbf{66.1} \\
\bottomrule
\end{tabular}
\end{minipage}
\hfill
\begin{minipage}[t]{0.50\textwidth}
\centering
\caption{\textbf{Pairwise win-rate} ablation of hierarchy, version, and phase correctness.}
\label{tab:model_comparison_b}
\begin{tabular}{l|l|c}
\toprule
Study & Model & Win Rate \\
\midrule
\multirow{2}{*}{Hierarchy} 
& SurgOnAir-base  & 39.4 \\
& SurgOnAir   & \textbf{60.6} \\
\midrule
\multirow{2}{*}{\makecell[l]{$\langle W \rangle$  \\ Formulation}} 
& SurgOnAir-v1 & 42.5 \\
& SurgOnAir & \textbf{57.5} \\
\midrule
\multirow{2}{*}{\makecell[l]{Phase \\ Correctness}} 
& SurgOnAir-base & 34.2 \\
& SurgOnAir & \textbf{65.8} \\
\bottomrule
\end{tabular}
\end{minipage}
\end{table}

\subsection{Narration Quality Evaluation}
We adopt Hulu-Med~\cite{jiang2025hulu} as a fixed comparison model, as its extensive medical pretraining has demonstrated strong effectiveness in surgical applications~\cite{jiang2025hulu}. Therefore, all pairwise evaluations are conducted between the evaluated model and Hulu-Med~\cite{jiang2025hulu}. 

\noindent
\textbf{Discussion.} Shown in Tab.~\ref{tab:model_comparison_a}, offline video-language models show limited effectiveness in surgical captioning. Despite having access to the entire video offline, LLaVA-Video-7B~\cite{lin2024video} and Qwen2.5-VL-7B~\cite{bai2025qwen3} achieve win-rate scores of only 11.3\% and 6.2\%, respectively, indicating a significant lack of domain-specific surgical understanding. In contrast, streaming models perform substantially better: LiveCC-7B~\cite{chen2025livecc} already surpasses all offline baselines with a score of 16.7\%. However, generic streaming pretraining remains insufficient for complex surgical scenarios. Training exclusively on in-domain surgical data using standard flat narration \textit{SurgOnAir-base} markedly improves the performance to 60.4\%. Building upon this, our final model \textit{SurgOnAir}, which incorporates hierarchical-aware modeling to capture both overarching procedural states and granular narrations, further elevates the score to 66.1\%. These results clearly demonstrate the crucial importance of both domain-specific data and hierarchical-structured supervision for accurate, real-time surgical narration.

\subsection{Temporal Alignment Evaluation}
We analyze clips from evaluation set which contains tool and action transitions. A prediction is temporally correct if the new tool or action appears in the generated narration within 1.5 seconds after the transition. Under this criterion, SurgOnAir correctly identifies \textbf{68.4\%} of clips, compared with \textbf{27.3\%} for HuluMed-7B\cite{jiang2025hulu}, which also tends to produce less specific narrations during transitions. These results suggest that hierarchy-aware modeling improves temporal adaptation by providing procedural context during streaming generation.

\subsection{Ablation Results}
For our ablation study, we perform pairwise comparisons to isolate the contribution of each designed component. The win rate reflects the relative preference between two competing variants via GPT-4o. Tab.~\ref{tab:model_comparison_b} presents a structured analysis focusing on three key aspects: hierarchy modeling, $\langle W \rangle$ formulations, and phase correctness.  

\noindent
\textbf{Hierarchy.} We evaluate the impact of hierarchical modeling by comparing our full model (\textit{SurgOnAir}) against \textit{SurgOnAir-base}, a baseline trained exclusively on flat narration from the surgical dataset without any hierarchical structure. The inclusion of our hierarchical design yields a substantial improvement, elevating the win rate from 39.4\% to 60.6\%. This gap indicates that hierarchical modeling provides essential structural guidance, enabling more coherent narratives compared to standard flat modeling.

\noindent
\textbf{$\langle W \rangle$ Formulations.} To validate our specific design for the state formulation, we compare our final model against an earlier variant, \textit{SurgOnAir-v1}. Unlike our proposed formulation that properly models state transitions, the v1 variant strictly predicts the state during every single word token generation step $\langle W_{t:t+k} \rangle = \langle S_t \rangle \oplus \langle N_{t:t+k} \rangle$. 
Forcing the model to generate state information for every short interval introduces excessive redundant overhead. Consequently, this modeling causes the model to over-rely on state prediction, inadvertently neglecting the most critical textual narration. Our \textit{SurgOnAir} model achieves a 57.5\% win rate against the 42.5\% of the \textit{SurgOnAir-v1} variant, confirming that a transition-aware $\langle W \rangle$ formulation is crucial for generating fluent and accurate narration.

\noindent
\textbf{Phase Correctness.} We study the specific contribution of accurate phase grounding. For this evaluation, we specifically select the test cases where \textit{SurgOnAir} correctly predicts the surgical phase, and compare its generated narrations against those produced by \textit{SurgOnAir-base} on the exact same videos. Under this strictly controlled setting, \textit{SurgOnAir} decisively outperforms the base model with a 65.8\% to 34.2\% win rate. This large margin explicitly demonstrates that successfully predicting the procedural phase provides strong, direct support for generating accurate and contextually aligned narration.

\begin{figure}[t]
    \centering
    \includegraphics[width=\linewidth]{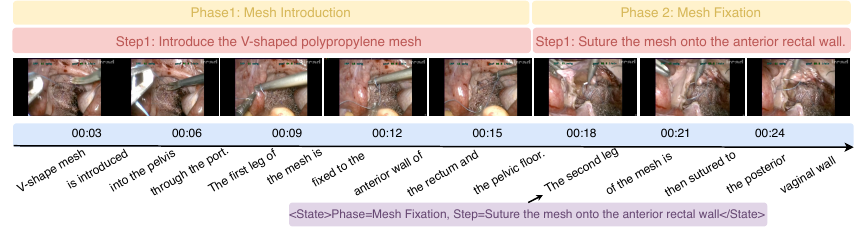}
    \caption{Qualitative result of SurgOnAir.}
    \label{fig:qualitative}
\end{figure}

\subsection{Qualitative Results}

Fig.~\ref{fig:qualitative} presents a qualitative example of our hierarchical streaming narration. Thanks to explicit hierarchical modeling, the narration transitions coherently across procedural boundaries. After detailing the actions within the \textit{Mesh Introduction} phase (e.g., "V-shape mesh is introduced into the pelvis..."), the model predicts the state transition at exactly 00:18. This structural anchor guides the continuous word stream into the subsequent \textit{Mesh Fixation} phase (e.g., "The second leg of the mesh is then sutured..."), ensuring temporal continuity. Beyond macro-level phase transitions, this hierarchical conditioning enables strict temporal alignment at the micro-level: the model recognizes and describes key instruments (e.g., the V-shaped mesh) and relevant anatomical structures (e.g., the pelvic floor) precisely at the moment of interaction. Such behavior demonstrates genuine temporally aligned narration, moving away from the limitations of isolated, frame-level captioning.

%% file: sections/4_Conclusion.tex
\section{Conclusion}
In this work, we present the first real-time surgical narration framework built upon hierarchical modeling. Our approach produces temporally aligned and hierarchical-aware descriptions, advancing real-time surgical understanding beyond conventional video captioning. This work lays an important foundation for future surgical automation and intelligent surgical teaching systems. Limitations include dataset scale and the absence of future action prediction, both of which offer promising directions for further exploration.

\subsubsection{\ackname} This work was supported in part by the Multi-Scale Medical Robotics Center, AIR@InnoHK, Hong Kong; in part by the NVIDIA Academic Grant Program.

\subsubsection{\discintname}
The authors have no competing interests to declare that are
relevant to the content of this article.